%% file: main.tex
\documentclass[10pt,twocolumn,letterpaper]{article}
\usepackage{cvpr}              
\usepackage{CJKutf8}
\usepackage{float}
\usepackage{amsmath}
\usepackage{amssymb}
\usepackage{booktabs}
\usepackage{algorithm}
\usepackage{multirow}
\usepackage{algorithmic}
\usepackage{graphicx}
\usepackage{bm}
\usepackage{colortbl}
\usepackage{appendix}
\usepackage{comment}
\usepackage{caption}

\input{preamble}

\definecolor{cvprblue}{rgb}{0.21,0.49,0.74}
\usepackage[pagebackref,breaklinks,colorlinks,citecolor=cvprblue]{hyperref}
\usepackage[capitalize]{cleveref}
\crefname{section}{Sec.}{Secs.}
\Crefname{section}{Section}{Sections}
\Crefname{table}{Table}{Tables}
\crefname{table}{Tab.}{Tabs.}
\def\paperID{6276} 
\def\confName{CVPR}
\def\confYear{2024}

\title{Semantic-aware SAM for Point-Prompted Instance Segmentation}
\author{Zhaoyang Wei$^{1}$\thanks{\ Equal contribution.}, Pengfei Chen$^{1*}$, Xuehui Yu$^{1*}$, Guorong Li$^{1}$, \\ 
Jianbin Jiao$^{1}$, Zhenjun Han$^{1}$\thanks{\ Corresponding authors. (hanzhj@ucas.ac.cn)} \\
{\small \textsuperscript{1}University of Chinese Academy of Sciences(UCAS)}}

\begin{document}
\maketitle



\begin{abstract}
Single-point annotation in visual tasks, with the goal of minimizing labelling costs, is becoming increasingly prominent in research.
Recently, visual foundation models, such as Segment Anything (SAM), have gained widespread usage due to their robust zero-shot capabilities and exceptional annotation performance. However, SAM's class-agnostic output and high confidence in local segmentation introduce \textbf{semantic ambiguity}, posing a challenge for precise category-specific segmentation. 
In this paper, we introduce a cost-effective category-specific segmenter using SAM.
To tackle this challenge, we have devised a Semantic-Aware Instance Segmentation Network (SAPNet) that integrates Multiple Instance Learning (MIL) with matching capability and SAM with point prompts. SAPNet strategically selects the most representative mask proposals generated by SAM to supervise segmentation, with a specific focus on object category information.
Moreover, we introduce the Point Distance Guidance and Box Mining Strategy to mitigate inherent challenges: \textbf{group} and \textbf{local} issues in weakly supervised segmentation. These strategies serve to further enhance the overall segmentation performance.
The experimental results on Pascal VOC and COCO demonstrate the promising performance of our proposed SAPNet, emphasizing its semantic matching capabilities and its potential to advance point-prompted instance segmentation.
The code is available at {\footnotesize\url{https://github.com/zhaoyangwei123/SAPNet}}.
\end{abstract}

\section{Introduction}
Instance segmentation seeks to discern pixel-level labels for both instances of interest and their semantic content in images, a crucial function in domains like autonomous driving, image editing, and human-computer interaction. Despite impressive results demonstrated by various studies \cite{DBLP:MASK-RCNN,DBLP:mask2former,DBLP:yolact,DBLP:solov2,DBLP:SOLO,DBLP:condinst} , the majority of these high-performing methods are trained in a fully supervised manner and heavily dependent on detailed pixel-level mask annotations, thereby incurring significant labeling costs. To address this challenge, researchers are increasingly focusing on weakly supervised instance segmentation, leveraging cost-effective supervision methods, such as bounding boxes \cite{DBLP:boxlevelset,DBLP:Boxinst,DBLP:discobox}, points \cite{DBLP:point2mask,DBLP:PSPS}, and image-level labels \cite{DBLP:IRNet,DBLP:BESTIE}.
\begin{figure}[tb!]
  \centering
    \includegraphics[height=0.505\linewidth]{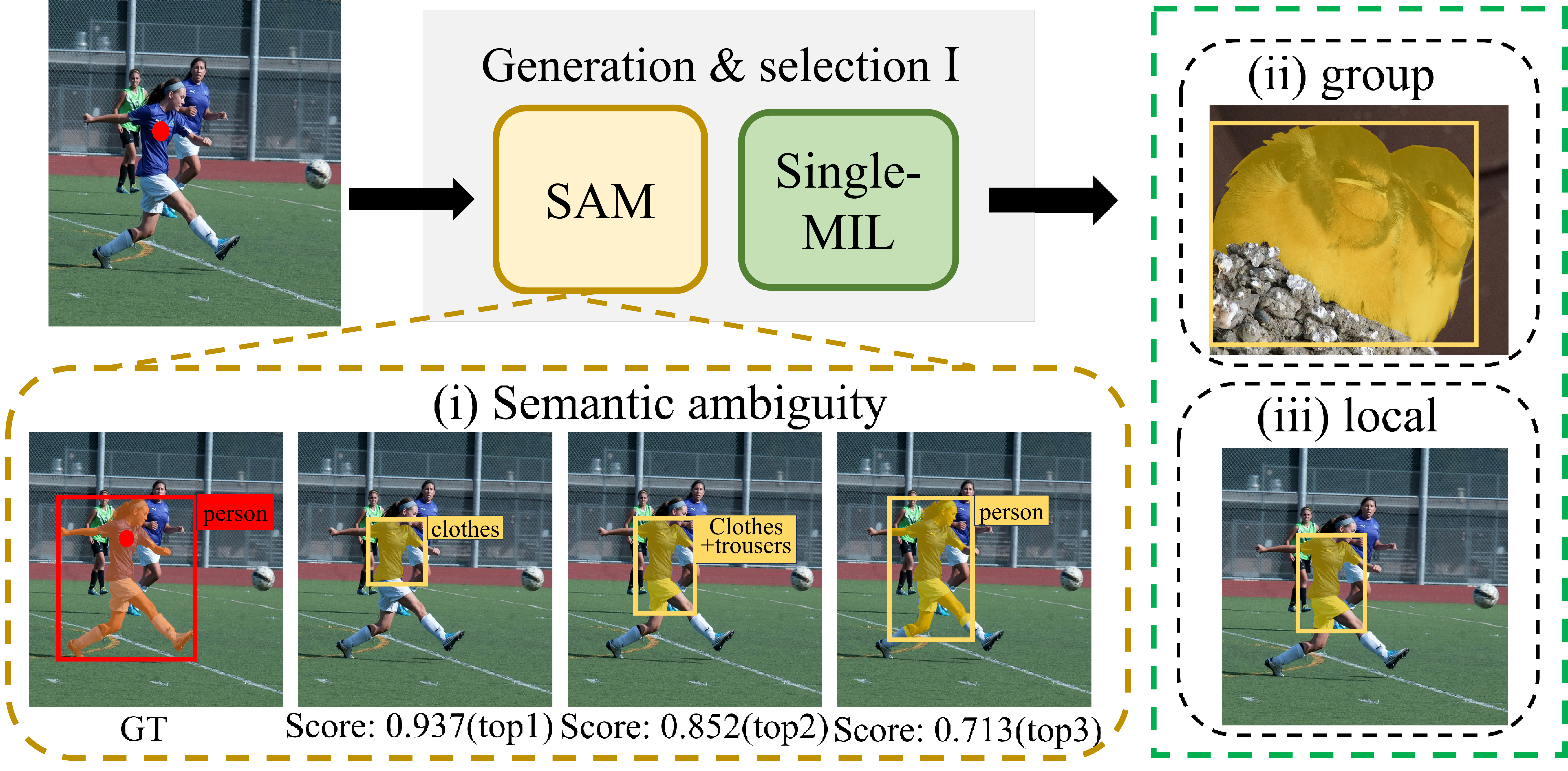}
\vspace{-15pt}
\setlength{\belowcaptionskip}{-0.5cm}
  \caption{Three Challenges Brought by SAM and single-MIL. Orange dash box illustrates that \textbf{semantic ambiguity} in SAM-generated masks, where it erroneously assigns higher scores to non-object categories like clothes, despite the person being our desired target. Green dash box depicts a comparison between mask proposals using single-MIL and SAPNet. It illustrates two primary challenges: \textbf{`group'}, where segmentation encounters difficulties in isolating individual targets among adjacent objects of the same category, and \textbf{`local'}, where MIL favors foreground-dominant regions, resulting in overlooked local details.}
\label{fig:problems}
\end{figure}

Recently, visual foundation models, such as Segment Anything (SAM)\cite{DBLP:segment-anything}, have been widely employed by researchers for their exceptional generalization capabilities and impressive annotation performance. Numerous studies based on SAM, such as \cite{DBLP:HQ-sam,DBLP:fast-sam} have emerged, building upon the foundations of SAM to further enhance its generalization capabilities and efficiency. However, these efforts have predominantly focused on improving the annotation performance of SAM. One limitation arises from SAM's lack of classification ability, resulting in class-agnostic segmentation results that fail to accurately segment specific categories as desired. 

To tackle the inherent semantic ambiguity in SAM and achieve specific-category segmentation, we propose integrating weak annotations with SAM, employing point annotations as prompts to imbue semantic information into SAM's outputs. A straightforward approach involves leveraging SAM's intrinsic scoring mechanism, selecting the top-scoring mask as the corresponding label for each category. However, when annotating object points are fed into the SAM, its category-agnostic characteristic tends to assign higher scores to parts of the object, resulting in generated mask annotations that fail to encompass the object as a whole. In Fig.~\ref{fig:problems} orange dashed box, we aim to obtain the `person' mask annotation, but SAM predicts the proposals of `clothes', `clothes+trousers' and 'person'. Relying solely on the score SAM provides is insufficient, as the highest score corresponds to `clothes' (col-2), which does not meet our specific needs.

To address this challenge, we have proposed SAPNet, a semantically-aware instance segmentation network designed for high-quality, end-to-end segmentation.
In this study, we design a proposal selection module (PSM) using the Multiple Instance Learning (MIL) paradigm to choose proposals that align closely with the specified semantic label. However, the MIL-based method relies on the classification score, often leading to group and local predictions \cite{DBLP:BESTIE,DBLP:WISE-Net,DBLP:wsddn}. In Fig.~\ref{fig:problems} green dashed box, the group issue is evident, where two objects of the same category are often both included when they are in close proximity. It also illustrates the local issue, where the MIL classifier frequently predicts the most discriminative region instead of the entire object. 
To overcome these limitations, we have introduced Point Distance Guidance (PDG) and Box Mining Strategies (BMS). Specifically, we penalize the selection results by calculating the Euclidean distances between the annotated points of identical categories enclosed within the proposals. Additionally, for more localized proposals, we filter out higher-quality proposals from their corresponding bags and dynamically merge them in scale.
By fully exploiting the positional clues to prevent local and group prediction, we aim to select the proposal that most effectively represents the object category in refinement stage. The primary contributions of this work can be outlined as follows:

1) We introduce SAPNet, an end-to-end semantic-aware instance segmentation network based on point prompts. SAPNet combines the visual foundation model SAM with semantic information to address its inherent semantic ambiguity, facilitating the generation of semantically-aware proposal masks.

2) We incorporate Point Distance Guidance (PDG) and Box Mining Strategies (BMS) to prevent local and group predictions induced by MIL-based classifiers in both the proposal selection and refinement stages.

3) SAPNet achieves state-of-the-art performance in Point-Prompted Instance Segmentation (PPIS), significantly bridging the gap between point-prompted and fully-supervised segmentation methods on two challenging benchmarks (COCO and VOC2012).

\section{Related Work}
\textbf{Weakly-Supervised Instance Segmentation (WSIS)} offers a practical approach for accurate object masks using minimal supervision. It spans a range of annotations, from image labels to bounding boxes. Research has focused on narrowing the performance gap between weakly and fully-supervised methods, primarily through box-level \cite{DBLP:Boxinst,DBLP:bbam,DBLP:BBTP} and image-level annotations \cite{DBLP:BESTIE,DBLP:inter-pixel}.
Box-based methods have explored structural constraints to guide the segmentation, as seen in BBTP \cite{DBLP:BBTP}, BoxInst \cite{DBLP:Boxinst}, and Box2Mask \cite{DBLP:box2mask}, and applied structural constraints to drive segmentation, treating it as a multiple-instance learning task or enforcing color consistency based on CondInst \cite{DBLP:condinst}.
These approaches, while innovative, can complicate training and sometimes neglect the object's overall shape due to their focus on local features and proposal generation, like MCG \cite{DBLP:MCG}. Conversely, the proposal-free methods, like IRN \cite{DBLP:inter-pixel}, rely on class relationships for mask production but can falter in accurately separating instances.
To preserve object integrity, recent methods such as Discobox \cite{DBLP:discobox} and BESTIE \cite{DBLP:BESTIE} integrate advanced semantic insights into instance segmentation using pairwise losses or saliency cues \cite{DBLP:Boxinst, DBLP:solov2, DBLP:label}. However, semantic drift remains an issue, with mislabeling or missed instances resulting in inferior pseudo labels \cite{DBLP:consistent} compromising segmentation quality.\\
\textbf{Pointly-Supervised Detection and Segmentation (PSDS)} cleverly balances minimal annotation costs with satisfactory localization accuracy. By introducing point annotations, WISE-Net \cite{DBLP:WISE-Net} , P2BNet \cite{DBLP:cpf}and BESTIE \cite{DBLP:BESTIE} improve upon weakly supervised methods that suffer from vague localizations. That only slightly increases the costs (by about 10\%) and is almost as quick as the image-level annotation, but that is far speedier than more detailed bounding box or mask annotations. Such precision allows for tackling semantic bias, as seen in methods like PointRend \cite{DBLP:pointrend}, which utilize multiple points for improved accuracy, despite requiring additional bounding box supervision.
Recent advancements in point-supervised instance segmentation, employed by WISE-Net and Point2Mask \cite{DBLP:point2mask}, show that even single-point annotations can yield precise mask proposals. WISE-Net skillfully localizes objects and selects masks, while BESTIE enhances accuracy using instance cues and self-correction to reduce semantic drift. Attnshift \cite{DBLP:Attnshift} advances this by extending single points to reconstruct entire objects.
Apart from their complexity, these methods have yet to fully demonstrate their effectiveness, indicating ongoing challenges in harnessing single-point annotations for image segmentation and presenting clear avenues for further research.\\
\textbf{Prompting and Foundation Models.} Prompt-based learning enables pretrained foundation models to adapt to various tasks using well-crafted prompts. SAM \cite{DBLP:segment-anything}, a prominent example in computer vision, exemplifies robust zero-shot generalization and interactive segmentation across multiple applications.
Additionally, SAM-based models like Fast-SAM \cite{DBLP:fast-sam} increases speed, HQ-SAM \cite{DBLP:HQ-sam} improves segmentation quality, and Semantic-SAM \cite{DBLP:semantic-sam} optimizes performance by training on diverse data granularities. Foundational models, pre-trained on large datasets, help improve generalization in downstream tasks, especially in data-scarce scenarios.
Basing on SAM, Rsprompter \cite{DBLP:rsprompter} utilizes SAM-derived pseudo labels for improved remote sensing segmentation, meanwhile, adaptations for medical imaging and video tracking are explored in A-SAM \cite{DBLP:a-sam} and Tracking Anything \cite{DBLP:tracking-anything}. Further, \cite{DBLP:sam-wsss1} and \cite{DBLP:sam-wsss2} have integrated SAM with Weakly Supervised Semantic Segmentation networks to refine pseudo labels. Our research builds upon these innovations, transforming point annotations into mask proposals in instance segmentation to significantly enhancing performance.\\
\begin{figure*}[tb!]
\begin{center}
    \begin{tabular}{ccc}
    \includegraphics[width=0.98\linewidth]{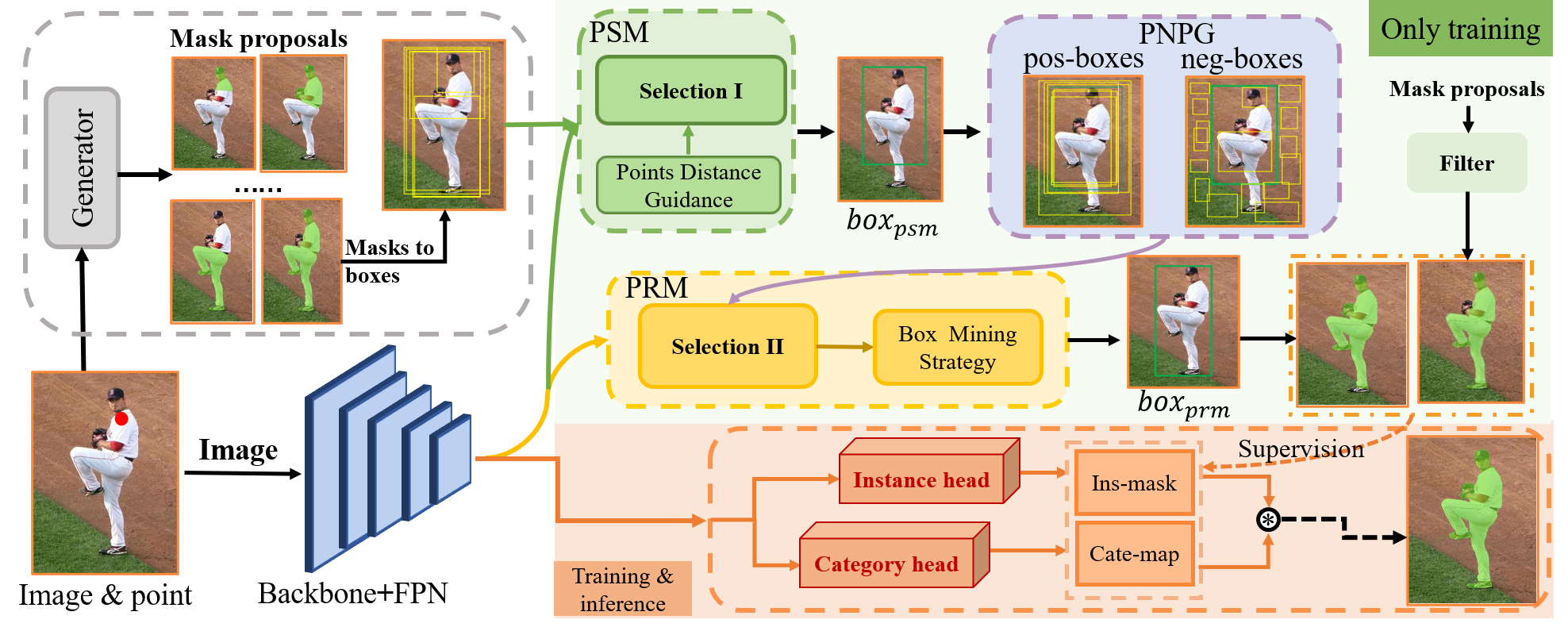}
    \end{tabular}
\vspace{-15pt}
\setlength{\belowcaptionskip}{-0.8cm}
   \caption{The framework of SAPNet comprises two components: one for generating mask proposals and another for their utilization in instance segmentation. The process starts with generating category-agnostic mask proposals using point prompts within a visual foundation model. That is followed by an initial proposal selection via MIL combined with PDG. Next, the PRM refines these proposals using positive and negative samples from PNPG, capturing global object semantics. Finally, augmented with the multi-mask proposal supervision, the segmentation branch aims to improve segmentation quality.
   }
\label{fig:framework}
\end{center}
\end{figure*}
\vspace{-15pt}
\section{Methodology}
\subsection{Overview}\label{sec:Overview}
The overview of our method is illustrated in Fig.~\ref{fig:framework}, SAPNet comprises of two branches: one dedicated to the selection and refinement of mask proposals to generate pseudo-labels and the other employing solov2 head~\cite{DBLP:solov2} for instance segmentation supervised by the generated pseudo labels.
The central focus of our approach is the pseudo-label generation branch, exclusively utilized during the training phase, which includes the PSM, PNPG, and PRM modules. Following the initial proposal inputs, the PSM employs multi-instance learning and a point-distance penalty to identify semantically rich proposals.
Subsequently, coupled with selected proposals from the PSM stage, the PNPG generates quality positive-negative bags to mitigate background and locality issues, emphasizing the primary regions of interest.
Then, the PRM processes these bags, which selects refined proposals from positive bags to improve final box quality. Ultimately, the mask mappings derived from these box proposals are utilized to guide the segmentation branch.
This guarantees the acquisition of high-quality category-specified mask proposals to supervise the segmentation branch.

\subsection{Proposal Selection Module}\label{sec:Prompt Selection Module}

SAM's limited semantic discernment causes category-agnostic labeling, leading to inconsistent proposal quality for the same objects. Employing these proposals directly for segmentation supervision could introduce noise and impair performance. Our goal is to design a category-specific segmenter, which needs to select the most semantically representative proposals for robust supervision.

Motivated by the insights from WSDDN \cite{DBLP:wsddn} and P2BNet \cite{DBLP:cpf}, our proposal selection module employs multi-instance learning and leverages labeling information to prioritize high-confidence proposals for segmentation. 
In the training phase, we leverage SAM\cite{DBLP:segment-anything} solely to generate category-agnostic proposals. To avoid excessive memory use and slow training, we convert them into box proposals using the minimum bounding rectangle, and combine with depth features \(F \in \mathbb{R}^{H \times W \times D}\) from the image \(I \in \mathbb{R}^{H \times W}\), serve as input to the PSM.
Utilizing our designed MIL loss, PSM precisely predicts each proposal's class and instance details. It selects the highest-scoring proposal as the semantically richest bounding box for each object, effectively choosing higher quality mask proposals.

\begin{figure}[!htb]
  \centering
    \includegraphics[height=0.315\linewidth]{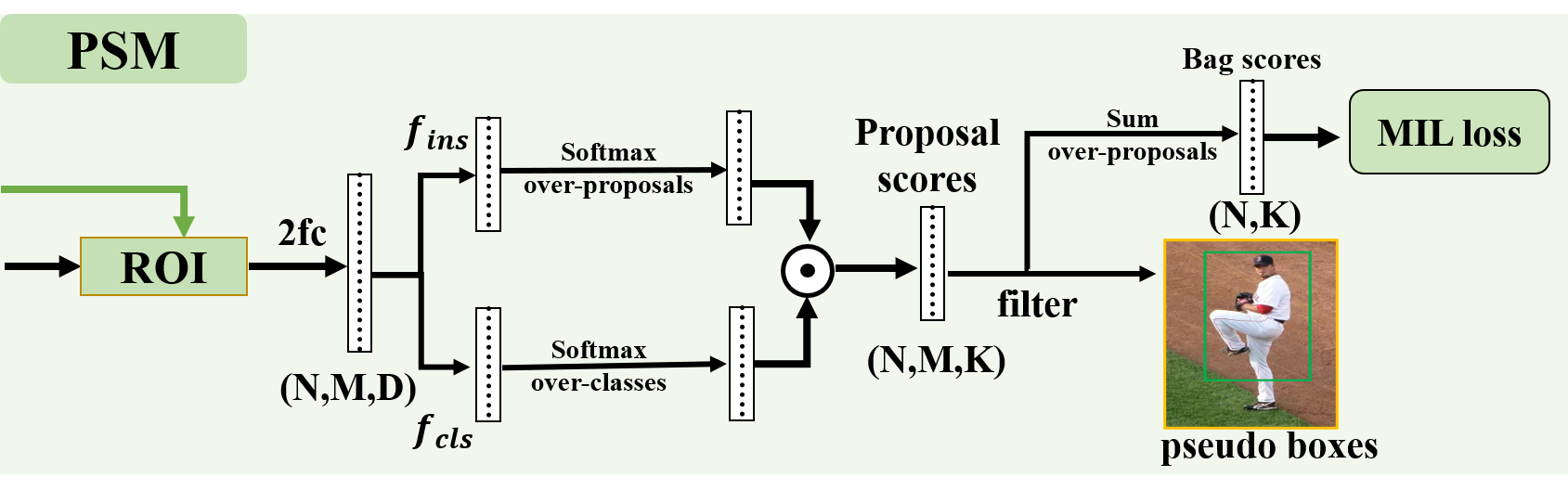}
    \vspace{-15pt}
    \caption{The mechanism of the proposal selection module.
    }
\label{fig:PSM}
\vspace{-20pt}
\end{figure}

Given an image \( I \) with \( N \) point annotations \( Y_n = \{ (p_i, c_i) \}_{i=1}^{N} \), where \( p_i \) is the coordinate of the annotated point and \( c_i \) is the class index. We transform each class-informative point \( p_i \) into \( M \) semantic mask proposals, which is further converted to a semantic proposal bag \( B_i \in \mathbb{R}^{M \times 4} \). As illustrated in Fig.~\ref{fig:framework}, after passing through a 7x7 RoIAlign layer and two fully-connected layers, features \( F_i \in \mathbb{R}^{M \times H \times W \times D}  \) are extracted from proposal bag \( B_i \).
Like in \cite{DBLP:wsddn} and \cite{DBLP:oicr}, the features \( F \) serve as input for the classification branch and instance branch, using fully-connected layer \( f \) and \( f' \) to generate \( \mathbf{W}_{{cls}}  \in \mathbb{R}^{M \times K} \) and \( \mathbf{W}_{ins}  \in \mathbb{R}^{M \times K} \). A softmax activation function over \( K \) class  and \( M \) instance dimensions yields the classification scores \( \mathbf{S}_{{cls}} \in \mathbb{R}^{M \times K} \) and instance scores \( \mathbf{S}_{{ins}} \in \mathbb{R}^{M \times K} \).
\begin{equation}\footnotesize
\vspace{-6pt}
\begin{aligned}
& \mathbf{W}_{cls} = f(\mathbf{F});
\ [\mathbf{S}_{cls}]_{mk} = e^{[\mathbf{W}_{cls}]_{mk}}\big/\sum\nolimits_{k=1}^{K} e^{[\mathbf{W}_{cls}]_{mk}}. \\
& \mathbf{W}_{ins} = f'(\mathbf{F}); 
[\mathbf{S}_{ins}]_{mk} = e^{[\mathbf{W}_{ins}]_{mk}}\big/\sum\nolimits_{m=1}^{M} e^{[\mathbf{W}_{ins}]_{mk}}.  \\
\label{Eq:initial score}
\end{aligned} 
\vspace{-10pt}
\end{equation}
where $[\cdot]_{mk}$ is the value in row $m$ and column $k$ of matrix.

\textbf{Point Distance Guidance.} SAM and MIL struggle with distinguishing adjacent objects of the same category, often merging two separate objects into one and giving high score. To combat this, we incorporate instance-level annotated point information and introduce a spatially aware selection with a point-distance penalty mechanism.

To address the challenge of overlapping objects and thereby enhance model optimization, we propose a strategy specifically aimed at penalizing instances of object overlap. For each m-th proposal within the set \(B_i\), we define \(t_{mj}=1\) to denote an overlap with any proposal in another identical class bag \(B_j\); otherwise, \(t_{mj}=0\). The penalty imposed increases in proportion to the distance of the overlapping objects from the proposal in question. This penalty, \(W_{dis}\), is represented using the Euclidean distance between the annotated points of the overlapping proposals. Subsequently, the reciprocal of \(W_{dis}\) is then passed through a sigmoid function to compute the distance score \(\mathbf{S}_{dis}\) for the proposal.

\vspace{-5pt}
\begin{equation}\footnotesize
\begin{aligned}
&[\mathbf{W}_{dis}]_{im}=\sum\limits_{j=1,j \neq i}^{N}\left\|p_i-p_j\right\| * t_{mj}.\\
&[\mathbf{S}_{dis}]_{im} = (1 / e^{-(1/[\mathbf{W}_{dis}]_{im})})^d.  \\
\label{Eq:MIL score}
\end{aligned} 
\vspace{-2pt}
\end{equation}
where $[\cdot]_{im}$ is the value at the row $i$ and column $m$ in the matrix, and $d$ is the exponential factor.

\textbf{PSM Loss.} 
The final score \( \mathbf{S} \) of each proposal is obtained by computing the Hadamard product of the classification score, the instance score, and the distance score, while the score \(\widehat{\mathbf{S}}\) for each proposal bag \( B_i \) is obtained by summing the scores of the proposals in \( B_i \). The MILloss of the PSM is constructed using the form of binary cross-entropy, and it is defined as follows:\\
\vspace{-7pt}
\begin{equation}\footnotesize
\begin{aligned}
& \mathbf{S}=\mathbf{S}_{cls} \odot \mathbf{S}_{ins} \odot \mathbf{S}_{dis} \in \mathbb{R}^{M \times K};
\widehat{\mathbf{S}}= \sum\limits_{m=1}^{M} [\mathbf{S}]_m \in \mathbb{R}^{K}. \\
& \mathcal{L}_{psm} = CE(\widehat{\mathbf{S}}, \mathbf{c}) =-\frac{\mathrm{1}}{N}\sum\limits_{n=1}^{N}\sum\limits_{k=1}^{K} \mathbf{c}_k \log(\widehat{\mathbf{S}}_k) + (1-\mathbf{c}_k)  \log(1-\widehat{\mathbf{S}}_k)
\label{Eq:bce loss}
\end{aligned} 
\vspace{-5pt}
\end{equation}
where $\mathbf{c} \in \{0, 1\}^{K}$ is the one-hot category's label. 

Utilizing the MILloss, the PSM module skillfully identifies each proposal's category and instance. The module selects the proposal with the highest score, marked as $ \mathbf{S} $, for a specific object and identifies a bounding box enriched with semantic information.

\subsection{Positive and Negative Proposals Generator}\label{sec:Positive and Negative Proposals Generator}
To further refine the selection of more accurate bounding boxes, we employ PNPG based on $box_{psm}$ selected via PSM. That consists of two components: PPG and NPG. The PPG is designed to generate a richer set of positive samples, enhancing bag's quality. Concurrently, the NPG is responsible for generating negative samples, which are crucial for assisting model training. These negative samples, including background samples for all objects and part samples for each, are crucial in resolving part issues and ensuring high-quality bounding box selection. The positive sample set \(B^+\) produced by PPG and the negative sample set \(\mathcal{U}\) generated by NPG are utilized for training the subsequent PRM.

\textbf{Positive Proposals Generator (PPG).} Within this phase, to implement adaptive sampling for the identified bounding box, we capitalize on the $box_{psm}$ derived from the PSM stage, coupled with the point distance penalty score \( \mathbf{S}_{{dis}} \) attributed to each proposal. To further elaborate, for each $box_{psm}$ (denoted as \( b_x^* \), \( b_y^* \), \( b_w^* \), \( b_h^* \)) isolated during the PSM phase, its dimensions are meticulously recalibrated leveraging a scale factor $v$ and its associated within-category inclusion score \( \mathbf{S}_{{dis}} \) to generate an augmented set of positive proposals \( (b_x, b_y, b_w, b_h) \). The formulation is defined as follows:
\begin{equation}\small
\setlength\abovedisplayskip{2pt}
\setlength\belowdisplayskip{0pt}
\begin{aligned}
b_w = (1 \pm v / \mathbf{S}_{dis}) \cdot b^*_w, \quad b_h = (1 \pm v / \mathbf{S}_{dis}) \cdot b^*_h, \\
b_x = b^*_x \pm (b_w - b^*_w)/2 , \quad b_y = b^*_y \pm (b_h - b^*_h)/2.
\setlength\abovedisplayskip{0pt}
\end{aligned}
\label{Eq: PRB sampling}
\end{equation} \\
These newly cultivated positive proposals are carefully integrated into the existing set $B_i$ to enhance the positive instances' pool. Such enhancements are pivotal in optimizing the training of the forthcoming PRM.

\textbf{Negative Proposals Generator(NPG).} MIL-based selection within a single positive bag may overemphasize the background noise, leading to inadequate focus on the object. To solve this, we create a negative bag from the background proposals post-positive bag training, which helps MIL maximize the attention towards the object.

Considering the image dimensions, we randomly sample proposals according to each image's width and height, for negative instance sampling. We assess the Intersection over Union (IoU) between these negatives and the positive sets, filtering out those below a threshold $T_{neg1}$.

Additionally, to rectify MIL localization errors, we enforce the sampling of smaller proposals with an IoU under a second threshold, $T_{neg2}$, from inside $box_{psm}$ based on its width and height, that is scored highest in PSM, as negative examples. These negative instances, partially capturing the object, drive the model to select high-quality bounding boxes that encompass the entire object. The PNPG is systematically elaborated upon in Algorithm\ref{Alg: positive and negative proposals generation (pnpg)}.
\begin{algorithm}[tb!]
{
\small
\caption{Positive and Negative Proposals Generation}
\label{Alg: positive and negative proposals generation (pnpg)}
\textbf{Input:} $T_{neg1}$,$T_{neg2}$,$box_{psm}$from PSM stage, image $I$, positive bags~$B^+$. \\
\textbf{Output:} Positive proposal bags $B^+$,Negative proposal set $\mathcal{U}$.\\
\vspace{-10pt}
\begin{algorithmic}[1]
\STATE \textit{// Step1: positive proposals sampling}
\FOR{$i \in N$,$N$ is the number of object in image $I$}
\STATE ${B^{+}_i} \leftarrow B_i$ ,$B_i \in B$;
\STATE ${B^{+}_i} = {B^{+}_i}\ \bigcup\ PPG(box_{psm}^i)$;
\ENDFOR
\STATE \textit{// Step2: background negative proposals sampling}
\STATE $\mathcal{U} \leftarrow \{\}$;
\STATE ${proposals} \leftarrow random\_sampling(I)$ for each image $I$; 
\STATE $iou = IOU(proposals, B_i)$ for each $B_i \in B$;
\IF{ $iou < T_{neg1}$ } 
 \STATE $\mathcal{U} = \mathcal{U}\ \bigcup\ proposals$;
\ENDIF
\STATE \textit{// Step3: part negative proposals sampling}
\FOR{$i \in N$,$N$ is the number of object in image $I$}
\STATE $proposals \leftarrow part\_neg\_sampling(box_{psm}^i)$ ;
\STATE $iou = IOU(proposals, box_{psm}^i)$ ;
\IF{ $iou < T_{neg2}$ } 
 \STATE $\mathcal{U} = \mathcal{U}\ \bigcup\ proposals$;
\ENDIF
\ENDFOR
\end{algorithmic}
}
\end{algorithm}
\subsection{Proposals Refinement Module}\label{sec:PRM}
In the PSM phase, we employ MIL to select high-quality proposals from bag $B^+$. However, as shown in Fig.~\ref{fig:framework}, the $box_{psm}$ outcomes derived solely from a single-stage MIL are suboptimal and localized. Inspired by PCL~\cite{DBLP:pcl}, we consider refining the proposals in a second phase. However, in contrast to most WSOD methods which choose to continue refining using classification information in subsequent stages, we have established high-quality positive and negative bags, and further combined both classification and instance branches to introduce the PRM module to refine the proposals, aiming to obtain a high-quality bounding box.

The PRM module, extending beyond the scope of PSM, focuses on both selection and refinement. It combines positive instances from the PPG with the initial set, forming an enriched $B^{+}$. Simultaneously, it incorporates the negative instance set $\mathcal{U}$ from NPG, providing a comprehensive foundation for PRM. This integration leads to a restructured MIL loss in PRM, replacing the conventional CELoss with Focal Loss for positive instances. The modified positive loss function is as follows:
\vspace{-7pt}
\begin{equation}
\begin{aligned}
\setlength\abovedisplayskip{0pt}
\setlength\belowdisplayskip{1pt}
\mathcal{L}_{pos}  = \frac{1}{N}\sum\limits_{i=1}^{N} \left< \mathbf{c}^{\mathrm{T}}_i, \mathbf{\widehat{S}}_i \right> \cdot {\rm FL}(\mathbf{\widehat{S}^{*}}_i, \mathbf{c}_i).
\label{Eq:L_{bag}}
\end{aligned}
\end{equation}
\noindent where ${\rm FL}$ is the focal loss~\cite{DBLP:retinanet_focalloss}, $\mathbf{\widehat{S}^*}_i$ and $\mathbf{\widehat{S}}_i$ represent the bag score predicted by PRM and PSM, respectively. ${\small{\left< \mathbf{c}^{\mathrm{T}}_i, \mathbf{\widehat{S}}_i \right>}}$ represents the inner product of the two vectors, meaning the predicted bag score of the ground-truth category. 

Enhancing background suppression, we use negative proposals and introduce a dedicated loss for these instances. Notably, these negative instances pass only through the classification branch for instance score computation, with their scores derived exclusively from classification. The specific formulation of this loss function is detailed below:

\vspace{-7pt}
\begin{equation}
\begin{aligned}
\setlength\abovedisplayskip{1pt}
\setlength\belowdisplayskip{1pt}
\beta =\frac{1}{N}\sum\limits_{i=1}^{N} \left< \mathbf{c}^{\mathrm{T}}_i, \mathbf{\widehat{S}}_i \right>,
\label{Eq:neg_pbr1}
\end{aligned}
\end{equation}
\begin{equation}
\begin{aligned}
\setlength\abovedisplayskip{1pt}
\setlength\belowdisplayskip{1pt}
\mathcal{L}_{neg} = - \frac{1}{\left|\mathcal{U}\right|}\sum\limits_{\mathcal{U}}\sum\limits_{k=1}^{K} \beta \cdot ([\mathbf{S}^{cls}_{neg}]_k)^{2} \log(1-[\mathbf{S}^{cls}_{neg}]_k).
\label{Eq:neg_pbr2}
\end{aligned}
\end{equation}
The PRM loss consists of the MIL loss $\mathcal{L}_{pos}$ for positive bags and negative loss $\mathcal{L}_{neg}$ for negative samples, \ie,
\begin{equation}
\begin{aligned}
\mathcal{L}_{prm} = \mathrm{\alpha} \mathcal{L}_{pos}  +(1-\mathrm{\alpha}) \mathcal{L}_{neg},
\label{Eq:PBR basic loss}
\end{aligned}
\end{equation}
\noindent where $\mathrm{\alpha}=0.25$ by default.

\textbf{Box Mining Strategy.} MIL's preference for segments with more foreground presence and SAM's tendency to capture only parts of an object often bring to final bounding boxes, $box_{prm}$, the `local' issue of MIL inadequately covers the instances. To improve the bounding box quality, we introduce a box mining strategy that adaptively expands $box_{select}$  from proposal selection in PRM, by merging it with the original proposals filter, aiming to address MIL's localization challenges.

The Box Mining Strategy (BMS) consists of two primary components:
(i) We select the top $ k $ proposals from the positive proposal bag $ B^+ $, to create a set $ G $. We evaluate the proposals in $ G $ against $ box_{select} $ based on IoU and size, using a threshold $ T_{min1} $. Proposals larger than $ box_{select} $ and with an IoU above $ T_{min1} $ undergo dynamic expansion through IoU consideration, which allows for the adaptive integration with $ box_{select} $. That mitigates the 'local' issue and maintains the bounding box's consistentcy to the object's true boundaries.
(ii) Frequently, issues related to locality can lead to an exceedingly low IoU between proposals and \(box_{select}\). Nonetheless, the ground truth box can fully encompass the $box_{part}$. Therefore, when component (i) conditions are unmet, if a proposal can entirely encapsulate \(box_{select}\), we reset the threshold \(T_{min2}\). Proposals surpassing this threshold adaptively merge with \(box_{select}\) to generate the final \(box_{prm}\),used to yield \(Mask_{prm}\). These two components collectively form our BMS strategy. A detailed procedure of this approach will be delineated in Algorithm2 of the supplementary materials.

\textbf{Loss Function.} After acquiring the final supervision masks, ${Mask}_{prm}$ and the filtered ${Mask}_{sam}$ in Multi-mask Proposals Supervision(MPS) in Sec. 7 of supplementary, we use them together to guide the dynamic segmentation branch. To comprehensively train SAPNet, we integrate the loss functions from the PSM and PRM, culminating in the formulation of the total loss for our model, denoted as $L_{total}$.
The aggregate loss function, $L_{total}$can be articulated as:
\begin{equation}\small
    \begin{aligned}
      \mathcal{L}_{total}=
      \mathcal{L}_{mask}
      + \mathcal{L}_{cls}
      + \lambda \cdot \mathcal{L}_{psm} 
      + \mathcal{L}_{prm} 
    \end{aligned}
\vspace{-3pt}
\end{equation}
\noindent where, $\mathcal{L}_{Dice}$ is the Dice Loss~\cite{DBLP:Diceloss}, $\mathcal{L}_{cls}$ is the Focal Loss\cite{DBLP:retinanet_focalloss}, and $\lambda$ is set as 0.25.

\section{Experiment}
\begin{table*}
\renewcommand\arraystretch{1.1}
\begin{center}
\resizebox{1.0\textwidth}{!}{
\begin{tabular}{l|c|c|c|c|ccc|ccc}
\specialrule{0.13em}{0pt}{0pt}  
Method & Ann.& Backbone & sched. &Arch. &$\rm mAP$ & $\rm mAP_{50}$ & $\rm mAP_{75}$ &$\rm mAP_{s}$ &$\rm mAP_{m}$ &$\rm mAP_{l}$\\
\specialrule{0.08em}{0pt}{0pt}  
\multicolumn{10}{c}{\textit{\textbf{Fully-supervised instance segmentation models.}}} \\

Mask R-CNN~\cite{DBLP:MASK-RCNN}&$\mathcal{M}$ &ResNet-50&1x&Mask R-CNN& 34.6 &56.5  &36.6 &18.3& 37.4& 47.2\\
YOLACT-700~\cite{DBLP:yolact}&$\mathcal{M}$ &ResNet-101&4.5x&YOLACT& 31.2 &54.0  &32.8 &12.1& 33.3& 47.\\
PolarMask~\cite{DBLP:MASK-RCNN}&$\mathcal{M}$ &ResNet-101&2x&PolarMask& 32.1 &53.7  &33.1 &14.7& 33.8& 45.3\\
SOLOv2~\cite{DBLP:solov2} &$\mathcal{M}$&ResNet-50&1x&SOLOv2&34.8& 54.9& 36.9&13.4& 37.8& 53.7\\
CondInst~\cite{DBLP:condinst} &$\mathcal{M}$&ResNet-50&1x&CondInst&35.3& 56.4& 37.4&18.0& 39.4& 50.4\\
SwinMR~\cite{DBLP:swin-transformer} & $\mathcal{M}$ &Swin-S&50e&SwinMR&43.2& 67.0 &46.1 &24.8& 46.3& 62.1\\
Mask2Former~\cite{DBLP:mask2former} &$\mathcal{M}$&Swin-S&50e&Mask2Former & 46.1&69.4&52.8 &25.4& 49.7& 68.5\\

\multicolumn{10}{c}{\textit{\textbf{Weakly-supervised instance segmentation models.}}} \\
IRNet~\cite{DBLP:IRNet} &$\mathcal{I}$& ResNet-50&1x&Mask R-CNN&6.1 &11.7   &5.5 &-&-&-\\
BESTIE~\cite{DBLP:BESTIE}  &$\mathcal{I}$  &  HRNet-48 &1x&Mask R-CNN& 14.3  & 28.0  & 13.2 &-&-&-\\
BBTP~\cite{DBLP:BBTP}& $\mathcal{B}$& ResNet-101&1x&Mask R-CNN& 21.1&45.5   &17.2 & 11.2 & 22.0 & 29.8\\
BoxInst~\cite{DBLP:Boxinst}& $\mathcal{B}$& ResNet-101&3x&CondInst& 33.2 &56.5&33.6 & 16.2 &35.3&45.1\\
DiscoBox~\cite{DBLP:discobox}& $\mathcal{B}$& ResNet-50&3x&SOLOv2& 32.0 &53.6&32.6 & 11.7 & 33.7 & 48.4 \\
Boxlevelset~\cite{DBLP:boxlevelset}& $\mathcal{B}$& ResNet-101&3x&SOLOv2& 33.4 &56.8&34.1 & 15.2 &36.8&46.8\\
WISE-Net~\cite{DBLP:WISE-Net}&$\mathcal{P}$ &ResNet-50&1x&Mask R-CNN&7.8  &18.2  & 8.8&-&-&-\\
BESTIE$^{\dagger}$~\cite{DBLP:BESTIE}&$\mathcal{P}$& HRNet-48 &1x&Mask R-CNN&17.7 & 34.0 &16.4 &-&-&-\\
AttnShift~\cite{DBLP:Attnshift}&$\mathcal{P}$ & Vit-B &50e&Mask R-CNN&21.2 &43.5&19.4 &-&-&-\\
\hline
\hline
 \rowcolor[rgb]{ 0.902,  0.902,  0.902} {SAM-SOLOv2}& $\mathcal{P}$& ResNet-50 &1x&SOLOv2& 24.6& 41.9  & 25.3 & 9.3& 28.6 & 38.1\\
 \rowcolor[rgb]{ 0.902,  0.902,  0.902} {MIL-SOLOv2}& $\mathcal{P}$& ResNet-50 &1x&SOLOv2& 26.8& 47.7  & 26.8 & 11.2 & 31.5 & 40.4\\
 \rowcolor[rgb]{ 0.902,  0.902,  0.902}\textbf{SAPNet(\emph{ours})}& $\mathcal{P}$& ResNet-50 &1x&SOLOv2& \textbf{31.2}& \textbf{51.8}  & \textbf{32.3} & \textbf{12.6} & \textbf{35.1} & \textbf{47.8}\\
 \rowcolor[rgb]{ 0.902,  0.902,  0.902}\textbf{SAPNet(\emph{ours})}$^{*}$& $\mathcal{P}$& ResNet-101&3x&SOLOv2&\textbf{34.6} &\textbf{56.0}  &\textbf{36.6} &\textbf{15.7} &\textbf{39.5}  &\textbf{52.1}  \\
\specialrule{0.13em}{0pt}{0pt}
\end{tabular}}
\end{center}
\vspace{-15pt}
\caption{Mask annotation($\mathcal{M}$), image annotation($\mathcal{I}$), box annotation($\mathcal{B}$) and point annotation($\mathcal{P}$) performance on COCO-17 val. `Ann.' is the type of the annotation and `sched.' means schedule. $^{*}$ is the multi-scale augment training for re-training segmentation methods, and other experiments are on single-scale training. SwinMR is Swin-Transformer-Mask R-CNN . SwinMR and Mask2Former use multi-scale data augment strategies for SOTA.}
\label{tab:coco_table1}
\vspace{-15pt}
\end{table*}
\subsection{Experimental Settings}
\textbf{Datasets.}
We use the publicly available MS COCO\cite{DBLP:coco} and VOC2012SBD \cite{DBLP:VOC} datasets for experiments. COCO17 has 118k training and 5k validation images with 80 common object categories. VOC consists of 20 categories and contains 10,582 images for model training and 1,449 validation images for evaluation.

\textbf{Evaluation Metric.} 
We use mean average precision mAP@[.5,.95] for the MS-COCO. The $\{ AP,AP_{50},AP_{75},AP_{Small},AP_{Middle},AP_{Large}$\} is reported for MS-COCO and for VOC12SBD segmentation, and we report AP$_{25, 50, 75}$.
The \(mIoU_{box}\) is the average IoU between predicted pseudo-boxes and GT-boxes in the training set. It measures SAPNet's ability to select mask proposals without using the segmentation branch.

\textbf{Implementation Details.} In our study, we employed the Stochastic Gradient Descent (SGD) optimizer, as detailed in \cite{DBLP:SGD}. Our experiments were conducted using the mmdetection toolbox \cite{DBLP:mmdetection}, following standard training protocols for each dataset. We used the ResNet architecture \cite{DBLP:resnet}, pretrained on ImageNet \cite{DBLP:imagenet}, as the backbone. For COCO, batch size was set at four images per GPU across eight GPUs, and for VOC2012, it was four GPUs. More details of the experiment are in Sec. 8 of the supplementary.
\subsection{Experimental Comparisons}
Tab.~\ref{tab:coco_table1} shows the comparison results between our method and previous SOTA approaches \cite{DBLP:MASK-RCNN,DBLP:solov2,DBLP:condinst,DBLP:mask2former,DBLP:swin-transformer} on COCO. 
In our experiments, we provide SAM with both the labeled points and the annotations generated by the point annotation enhancer \cite{DBLP:cpf}. SAM then utilizes these inputs to generate subsequent mask proposals for selection and supervision. For fair comparison, we design two baselines: the top-1 scored mask from SAM and MIL-selected SAM mask proposals are used as SOLOv2 supervision, respectively. Tab.~\ref{tab:coco_table1} shows our method substantially surpasses these baselines in performance.

\textbf{Comparison with point-annotated methods.} Our approach achieves a 31.2 $AP$ performance with a ResNet-50 backbone, surpassing all previous point-annotated methods, including BESTIE on HRNet-48 and AttnShift on Vit-B.
Our model exhibits significant improvements under a 1x training schedule, with a 13.5 $AP$ increase
when compared to the previous SOTA method, BESTIE.
Furthermore, under a 3x training schedule, SAPNet outperforms AttnShift, which relies on large model training, with 13.4 $AP$, 
improvements. Importantly, our method is trained end-to-end without needing post-processing, achieving SOTA performance in point-annotated instance segmentation.

\textbf{Comparison with other annotation-based methods.}  Our SAPNet has significantly elevated point annotation, regardless of point annotation's limitations in annotation time and quality compared to box annotation. Utilizing a ResNet-101 backbone and a 3x training schedule, SAPNet surpasses most box-annotated instance segmentation methods, achieving a 1.4 $AP$ improvement over BoxInst. Moreover, SAPNet's segmentation performance nearly matches the mask-annotated methods, effectively bridging the gap between point-annotated and these techniques.

\textbf{Segmentation performance on VOC2012SBD.} Tab.~\ref{tab:voc_12_seg} compares segmentation methods under different supervisions on the VOC2012 dataset. SAPNet reports an enhancement of 7.7 $AP$ over the AttnShift approach, evidencing a notable advancement in performance. Thereby, it significantly outstrips image-level supervised segmentation methods. Additionally, SAPNet surpasses box-annotated segmentation methods, such as BoxInst by 3.4 $AP_{50}$ and DiscoBox by 2.6 $AP_{50}$. Further, our point-prompted method achieves $92.3\%$ of the Mask-R-CNN.
\begin{table}[tb!]
    \centering
    \resizebox{1.\linewidth}{!}{
    \begin{tabular}{l|ccccc}
    \specialrule{0.13em}{0pt}{0pt}
         Method&Sup.&Backbone&$ AP_{25}$&$ AP_{50}$&$ AP_{75}$\\
    \specialrule{0.08em}{0pt}{0pt} 
        Mask R-CNN~\cite{DBLP:MASK-RCNN}&$\mathcal{M}$&R-50&78.0&68.8&43.3\\
        Mask R-CNN~\cite{DBLP:MASK-RCNN}&$\mathcal{M}$&R-101&79.6&70.2&45.3\\
        
        BoxInst~\cite{DBLP:Boxinst} &$\mathcal{B}$&R-101&-&61.4&37.0\\
        DiscoBox &$\mathcal{B}$&R-101&72.8& 62.2 &37.5 \\
        BESTIE~\cite{DBLP:BESTIE}& $\mathcal{I}$ &HRNet &53.5&41.7&24.2 \\
        IRNet~\cite{DBLP:IRNet} &$\mathcal{I}$&R-50&-&46.7&23.5\\
        BESTIE$^{\dagger}$~\cite{DBLP:BESTIE}& $\mathcal{I}$ &HRNet &61.2&51.0&26.6 \\
         WISE-Net~\cite{DBLP:WISE-Net}&$\mathcal{P}$&R-50&53.5&43.0&25.9 \\
         BESTIE~\cite{DBLP:BESTIE}&$\mathcal{P}$&HRNet&58.6&46.7&26.3 \\
         BESTIE$^{\dagger}$~\cite{DBLP:BESTIE}&$\mathcal{P}$&HRNet&66.4&56.1&30.2 \\
         Attnshift~\cite{DBLP:Attnshift} &$\mathcal{P}$&Vit-S&68.3 &54.4&25.4 \\
         Attnshift$^{\dagger}$~\cite{DBLP:Attnshift} &$\mathcal{P}$&Vit-S&70.3 &57.1&30.4 \\         

         \hline
         \hline
         \rowcolor[rgb]{ 0.902,  0.902,  0.902}\textbf{SAPNet(\emph{ours})}&$\mathcal{P}$&R-101 &\textbf{76.5}&\textbf{64.8}&\textbf{40.8} \\
             \specialrule{0.13em}{0pt}{0pt}
    \end{tabular}}
    \vspace{-8pt}
    \caption{Instance segmentation performance on the VOC2012 $test$ set. $^{\dagger}$ indicates applying MRCNN refinement.}
    \vspace{-25pt}
    \label{tab:voc_12_seg}
\end{table}
\subsection{Ablation Studies}
More experiments have been conducted on COCO to further analyze SAPNet's effectiveness and robustness.

\begin{figure}[!htb]
  \centering
    \includegraphics[height=0.335\linewidth]{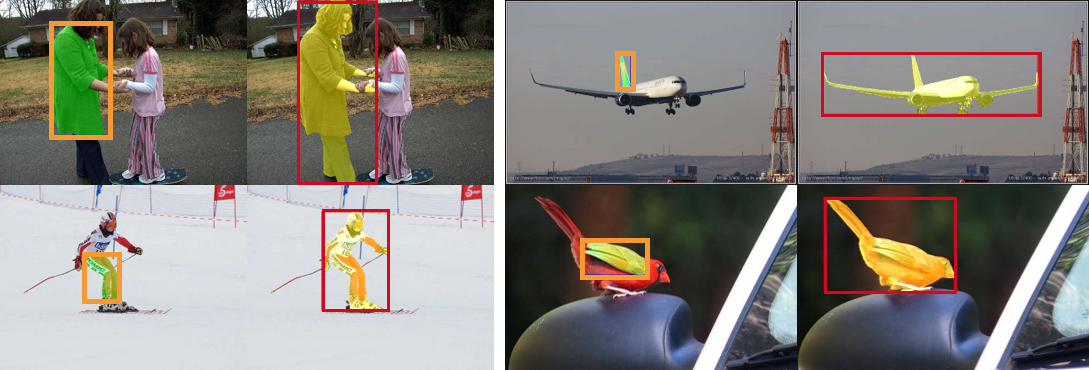}
    \vspace{-5pt}
    \caption{The comparative visualization between SAM-top1 and SAPNet is presented, showcasing SAM's segmentation outcomes in green masks and our results in yellow. The orange and red bounding boxes highlight the respective mask boundaries.
    }
\label{fig:sam-our}
\vspace{-10pt}
\end{figure}

\textbf{Training Stage in SAPNet.}  
The ablation study of the training stage is given in Tab.~\ref{tab:training stage}. 
We trained solov2 using the top-1 scored mask provided by SAM and compared it to the two training strategies of SAPNet. In the two-stage approach, the segmentation branch and multiple-mask supervision of SAPNet are removed. Instead, we use the selected mask to train a standalone instance segmentation model, as described by \cite{DBLP:solov2}. The end-to-end training method corresponds to the architecture illustrated in Fig.~\ref{fig:framework}. Our findings indicate that our method is more competitive than directly employing SAM (31.2 $AP$ \emph{vs} 24.6 $AP$), and the visualization of Fig.~\ref{fig:sam-our} shows us this enhancement. Moreover, the end-to-end training strategy boasts a more elegant model structure and outperforms the two-stage approach in overall efficiency (31.2 $AP$ \emph{vs} 30.18 $AP$).
\begin{table}[tb!]
\begin{center}
    \resizebox{0.48\textwidth}{!}{
\begin{tabular}{c|c|ccc}
\specialrule{0.13em}{0pt}{0pt}
train stage on coco & sched. & \(AP\) & \(AP_{50}\)& \(AP_{75}\)                \\
\specialrule{0.08em}{0pt}{0pt}
SAM-top1  &1x & 24.6 &   41.9   & 25.3 \\
Two stage  &1x & 30.2 &   49.8   & 31.5 \\

End to end & 1x       & 31.2 & 51.8 & 32.3 \\
\specialrule{0.13em}{0pt}{0pt}
\end{tabular}}
\end{center}
\vspace{-15pt}
\caption{The experimental comparisons of segmenters in COCO dataset, SAM-top1 is the highest scoring mask generated by SAM.}
\vspace{-15pt}
\label{tab:training stage}
\end{table}

\textbf{Effect of Each Component.}
Given the limited performance of SAM-top1, we opted for the single-MIL as our baseline. With a preliminary selection using MIL1, we have achieved a segmentation performance of 26.8 $AP$. 
\textbf{i) Point Distance Guidance.} We updated the proposal scores from the existing MIL by integrating the PDG module into the foundational MIL selection. This approach successfully segments adjacent objects of the same category, improving the segmentation performance by 0.7 points (27.5 vs 26.8). 
\textbf{ii) MIL2.} Building on the previous step, we incorporate a second MIL selection module to refine the initially selected boxes, resulting in a performance increment of 0.2 points.
\textbf{iii) PNPG.} For MIL2, we devised the positive-negative sample sets, aiming to enhance the input quality for the PRM module and use the negative samples to suppress background. This adjustment leads to a segmentation performance boost of 2 points (29.7 vs 27.7).
\textbf{iv) BMS.} Within the PRM, we refine the selected boxes using BMS, pushing the segmentation performance up by 1.1 points (30.8 vs 29.7).
\textbf{v) MPS.} Utilizing MPS for segmentation branch supervision yields a 0.4-point performance improvement.
\begin{table}[tb!]
\begin{center}
\begin{tabular}{cccccc|c}
\specialrule{0.13em}{0pt}{0pt}
mil1 & PDG & mil2 & PNPG & BMS & MPS &$mAP$ \\
\specialrule{0.08em}{0pt}{0pt}
 \checkmark&            &            &            &   &  &26.8  \\
 \checkmark& \checkmark &            &            &   &  &27.5  \\
 \checkmark& \checkmark & \checkmark &            &   &  &27.7 \\
 \checkmark& \checkmark & \checkmark & \checkmark &   &  &29.7 \\
 \checkmark& \checkmark & \checkmark & \checkmark &\checkmark&  &30.8 \\
 \checkmark& \checkmark & \checkmark & \checkmark &\checkmark&\checkmark&31.2 \\
\specialrule{0.13em}{0pt}{0pt}
\end{tabular}
\end{center}
\vspace{-15pt}
\setlength{\belowcaptionskip}{-0.5cm}
\caption{The effect of each component in SAPNet: proposal selection module(MIL1), point distance guidance(PDG), positive and negative proposals generator(PNPG), proposal selection module(MIL2), box mining strategy(BMS), and Multi-mask Proposals Supervision(MPS) in Sec. 7 of  supplementary.}\vspace{-5pt}
\label{tab:each component}
\end{table}

\textbf{Threshold of BMS.} 
For point refinement, there are two constraints (described in Sec.~\ref{sec:PRM}).
$T_{min1}$ and $T_{min2}$ are thresholds of the Box Mining Strategy. 
In Tab.~\ref{tab:BMS threshold}, it shows that the two constraints together to obtain performance gain. After multiple experiments, we have found that there is a significant performance improvement when $T_{min1}$ and $T_{min2}$ are set to 0.6 and 0.3, respectively.

\textbf{Components of PNPG.}
Tab.~\ref{tab:pnpg} presents the results of a dissected ablation study on the Positive and Negative Proposals Generator(PNPG), illustrating the respective impacts of the positive and negative examples on the model's performance. It is evident that the construction of negative examples plays a significant role in enhancing model efficacy. Furthermore, the beneficial effects of both positive and negative examples are observed to be cumulative.

\textbf{Performance Analysis.} 
As presented in Tab.~\ref{tab:gap}, we conducted a statistical analysis to validate SAPNet's capability to address 'local' issue and compare the outcomes selected by the single-MIL with those obtained by SAPNet in the absence of segmentation branch integration. Specifically, the part problem generated by the single-MIL, where MIL is inclined to select proposals with a higher proportion of foreground, is exemplified in Fig. 6 of  supplementary. On this premise, we initially establish an evaluative criterion \( R_v = \frac{area_{mask}}{area_{box}} \), which is the ratio of the mask area to the bounding box area. Subsequently, we compute \( R_{v_i} \) for each proposal within the proposal bag corresponding to every instance across the entire COCO dataset and select the maximum \( R_{v_{max}} \) to compute the mean value over the dataset, which is then designated as the threshold \( T_{rv} \). Ultimately, we identify the ground truth \( R_{v_{gt}} \) and objects where \( R_{v_{max}} \) exceeds \( T_{rv} \) and calculates the discrepancy between \( R_v \) values selected by single-MIL and SAPNet. The description is as follows:
\vspace{5pt}
\begin{equation}\small
\setlength\abovedisplayskip{2pt}
\setlength\belowdisplayskip{0pt}
\begin{aligned}
Gap_{single} = Rv_{single} - Rv_{gt}, \quad Gap_{our} = Rv_{our} - Rv_{gt}.
\setlength\abovedisplayskip{0pt}
\end{aligned}
\label{Eq: area}
\end{equation} 
\vspace{-10pt}

Tab.~\ref{tab:gap} shows that the proposed SAPNet mitigates the locality issue faced by the single-MIL. Furthermore, the boxes selected via SAPNet exhibit a substantially higher IoU with GT than those selected by the single-MIL.
\begin{table}[tb!]
\begin{center}
    \resizebox{0.48\textwidth}{!}{
\begin{tabular}{c|c|cccccc}
\specialrule{0.13em}{0pt}{0pt}
$T_{min1}$ & $T_{min2}$ & \(AP\) & \(AP_{50}\)& \(AP_{75}\) & \(AP^s\) & \(AP^m\) & \(AP^l\)              \\
\specialrule{0.08em}{0pt}{0pt}
0.5  &0.3 & 30.9 &   51.3   & 32.0 & 12.2 &   34.7   & 47.4\\

0.5  &0.4 &30.7 &   51.2   & 31.8 & 11.9 &   34.7   & 47.1\\

0.6  &0.3 & \textbf{31.2} &   \textbf{51.8}   & \textbf{32.3} & \textbf{12.6} &   \textbf{35.1}   & \textbf{47.8}\\

0.6  &0.4 & 30.8 &   51.1   & 32.0 & 12.1 &   34.7   & 47.3\\

0.7  &0.3 & 31.0 &   51.5   & 32.2 & 12.6 &   34.9   & 47.3\\

0.7  &0.4 & 30.7 &   51.1   & 31.9 & 12.0 &   34.6   & 47.2\\

\specialrule{0.13em}{0pt}{0pt}
\end{tabular}}
\end{center}
\vspace{-15pt}
\caption{ Constraints in box mining strategy.}
\vspace{-10pt}
\label{tab:BMS threshold}
\end{table}

\begin{table}[tb!]
\begin{center}
\begin{tabular}{c|c|ccc}
\specialrule{0.13em}{0pt}{0pt}
\multicolumn{2}{c|}{\bf PNPG} & \multirow{2}{*}{\bf $AP$} & \multirow{2}{*}{\bf $AP_{50}$} & \multirow{2}{*}{$AP_{75}$} \\
\cline{1-2}
\(PPG\) & \(NPG\) &  &  &  \\
\specialrule{0.08em}{0pt}{0pt}
 &  & 29.3 & 49.7 & 30.0\\
\checkmark &  & 29.8 & 50.5 & 30.8\\
  & \checkmark &  30.7 &51.2 &31.7 \\
\checkmark  & \checkmark & \textbf{31.2} & \textbf{51.8} & \textbf{32.3} \\
\specialrule{0.13em}{0pt}{0pt}
\end{tabular}
\end{center}
\vspace{-15pt}
\caption{Meticulous ablation experiments in PNPG}
\label{tab:pnpg}
\vspace{-10pt}
\end{table}

\begin{table}[tb!]
\begin{center}
\begin{tabular}{l|c|c}
\specialrule{0.13em}{0pt}{0pt}
Method & $Gap$ & $mIoU_{box}$ \\
\specialrule{0.08em}{0pt}{0pt}
Single-MIL  & 0.199  & 63.8 \\
SAPNet  & 0.131 & 69.1 \\
\specialrule{0.13em}{0pt}{0pt}
\end{tabular}
\end{center}
\vspace{-15pt}
\caption{Experimental analysis with part problem.}
\label{tab:gap}
\vspace{-15pt}
\end{table}
\vspace{-2pt}
\section{Conclusion}
In this paper, we propose SAPNet, an innovative end-to-end point-prompted instance segmentation framework. SAPNet transforms point annotations into category-agnostic mask proposals and employs dual selection branches to elect the most semantic mask for each object, guiding the segmentation process. 
To address challenges such as indistinguishable adjacent objects of the same class and MIL's locality bias, we integrate PDG and PNPG, complemented by a Box Mining Strategy for enhanced proposal refinement. SAPNet uniquely merges segmentation and selection branches under multi-mask supervision, significantly enhancing its segmentation performance. Extensive experimental comparisons on VOC and COCO datasets validate the SAPNet's effectiveness in point-prompted instance segmentation.
\footnotesize\section{Acknowledgements}
\vspace{-5pt}
This work was supported in part by the Youth Innovation Promotion Association CAS, the National Natural Science Foundation of China (NSFC) under Grant No. 61836012, 61771447 and 62272438, and the Strategic Priority Research Program of the Chinese Academy of
Sciences under Grant No.XDA27000000.

{
    \small
    \bibliographystyle{ieeenat_fullname}
    \bibliography{main}
}

\input{supplementary}

\end{document}

%% file: preamble.tex
\usepackage[dvipsnames]{xcolor}


%% file: supplementary.tex
\clearpage
\setcounter{page}{1}
\maketitlesupplementary

%
%
\section{Segmentation branch}
\label{sec:MPS}
\textbf{Multi-mask Proposals Supervision.} We utilize SOLOv2\cite{DBLP:solov2} as our segmentation branch for its efficiency. Alongside using $Mask_{prm}$ from the $box_{prm}$ mapping for segmentation supervision, we integrated a proposal filter, which ranks proposals in $H \in \mathbb{R}^{N \times M}$ based on PRM scores, extracting initial masks to yield $Mask_{sam}$. Combined with $Mask_{prm}$, these guide the segmentation network. The loss function is as follows:
\vspace{-3pt}
\begin{equation}\small
    \begin{aligned}
      \mathcal{L}_{mask}= 
       \mathcal{L}_{Dice}(Mask_{pre},Mask_{prm}) \\
      + \gamma \cdot \mathcal{L}_{Dice}(Mask_{pre},Mask_{sam})
    \end{aligned}
\end{equation}
\noindent where $\mathcal{L}_{Dice}$ is the Dice Loss~\cite{DBLP:Diceloss}, $\gamma$ is set as 0.25.

\textbf{Inference.} 
For SAPNet's inference process, only the segmentation branch is retained after training, which is identical to the original instance segmentation model\cite{DBLP:solov2}. Given an input image, mask predictions are directly produced via an efficient Matrix-NMS technique. The pseudo-mask selection procedures of PSM, PRM, and other MIL-based modules only introduce computational overhead during training; they are entirely cost-free during inference.
\section{Visualization and ablation studies on COCO}
\label{sec:visual}
\textbf{Implementation Details.} On COCO and VOC2012SBD datasets, the initial learning rates were 1.5 × $10^{-2}$ and 2 × $10^{-3}$, respectively, reduced by a factor of 10 at the 8th and 10th epochs. In PDG, the exponential factor \(d\) was set to 0.015. For NPG, the IoU threshold $T_{neg1}$ and $T_{neg2}$ were 0.3 and 0.5, respectively. For BMS, the $k$ was set to 3. From the mask bag, we selected masks with the highest, medium, and lowest scores outside of $Mask_{prm}$ for MPS to accelerate the convergence of segmentation. Training spanned 12 epochs. Single-scale evaluation (1333 × 800) was used for the 1x schedule. For the 3x schedule, a multi-scale training approach was adopted, with the image's short side resized between 640 and 800 pixels (in 32-pixel increments) and MPS will be removed . Inference was conducted using single-scale evaluation.

\begin{figure*}[ht]
\setlength{\abovecaptionskip}{0cm} 
\setlength{\belowcaptionskip}{0cm} 
\begin{center}
    \begin{tabular}{ccc}
    \includegraphics[width=0.98\linewidth]{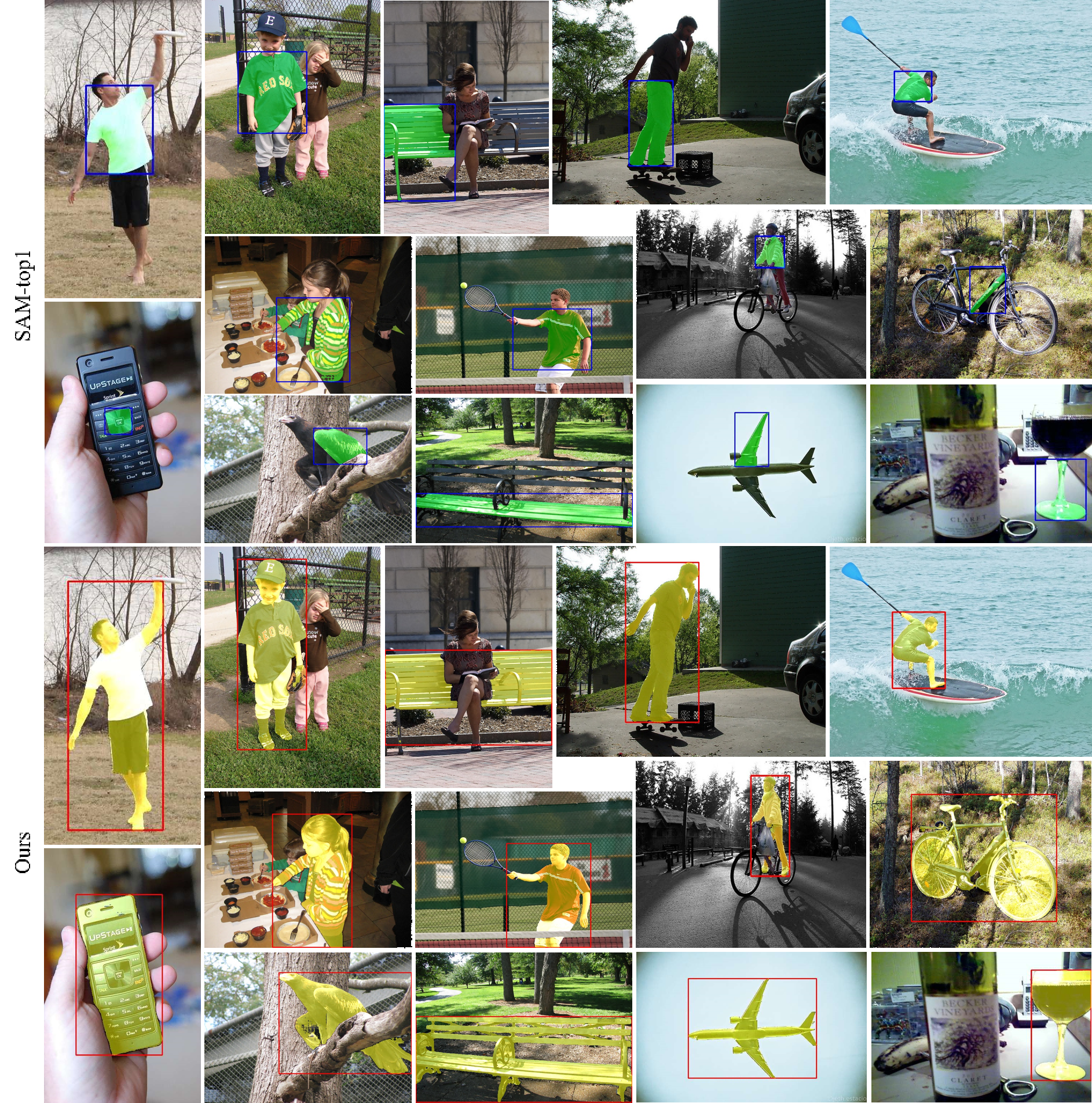}
    \end{tabular}
   \captionsetup{width=0.98\linewidth}
   \caption{Visualization comparison between SAM-top1 and SAPNet on COCO 2017 dataset about semantic ambiguity, showing SAM's segmentation outcomes top-1 in green masks and our results in yellow masks. The blue and red bounding boxes highlight the respective mask boundaries.
   }
\label{fig:visual1}
\end{center}
\end{figure*}

\textbf{Visualization.} The proposed SAPNet significantly mitigates SAM's semantic ambiguity. Fig.~\ref{fig:visual1} reveals SAM's limitations in discerning semantic significance from point prompts (the category indicated by each point). With SAPNet's refinement, each object is equipped with a mask that accurately represents its semantic category. The fidelity of masks selected by SAPNet is distinctly higher than that of top-1 masks from SAM.

\begin{figure*}[ht]
\setlength{\abovecaptionskip}{0cm} 
\setlength{\belowcaptionskip}{0cm} 
\begin{center}
    \begin{tabular}{ccc}
    \includegraphics[width=0.98\linewidth]{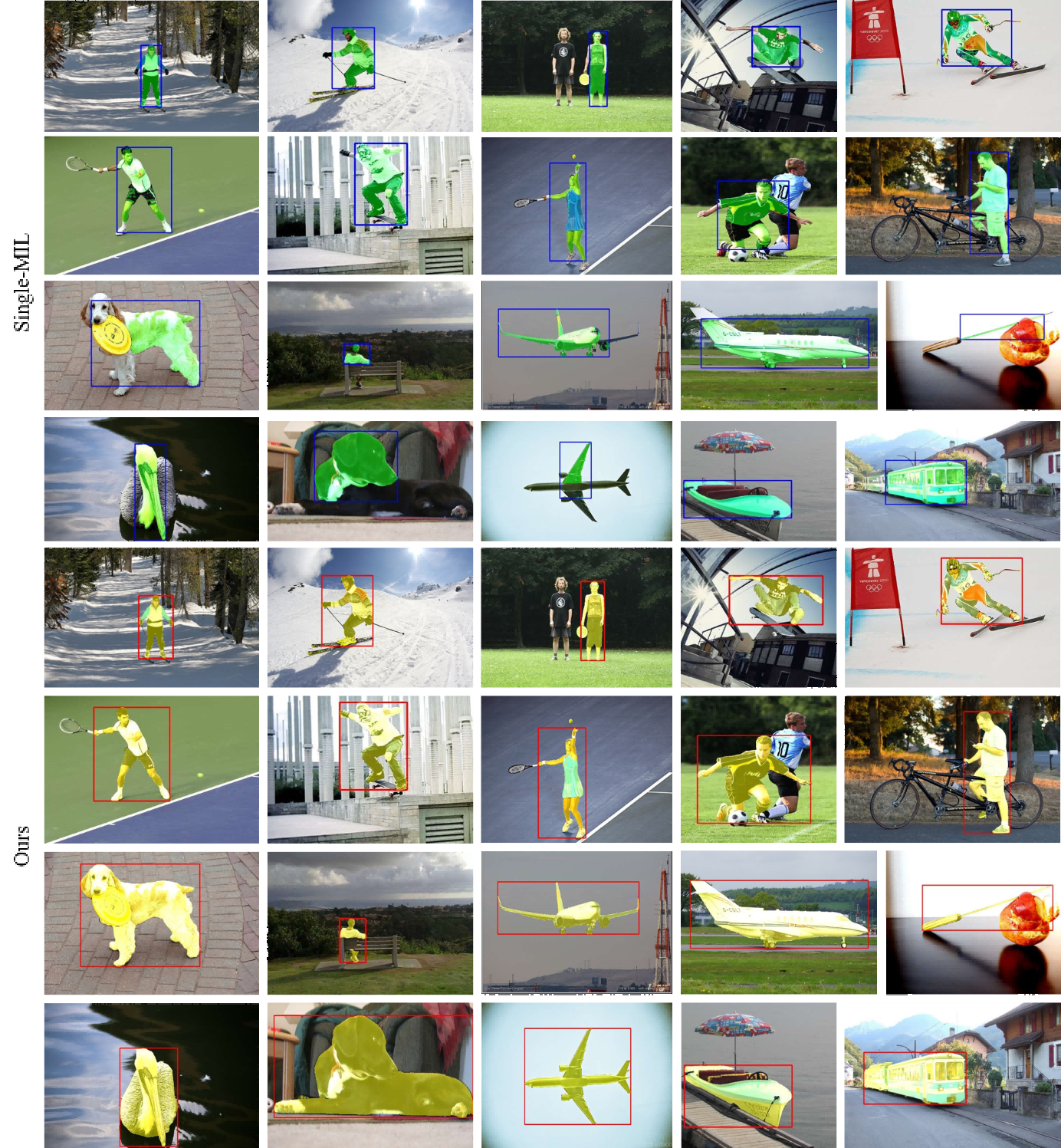}
    \end{tabular}
   \captionsetup{width=0.98\linewidth}
   \caption{Visualization comparison between the single-MIL and SAPNet on COCO 2017 dataset about local segmentation, showing single-MIL's outcomes in green masks and our results in yellow masks. The blue and red bounding boxes highlight the respective mask boundaries.
   }
\label{fig:visual0}
\end{center}
\end{figure*}

\begin{figure*}[ht]
\setlength{\abovecaptionskip}{0cm} 
\setlength{\belowcaptionskip}{0cm} 
\begin{center}
    \begin{tabular}{ccc}
    \includegraphics[width=0.98\linewidth]{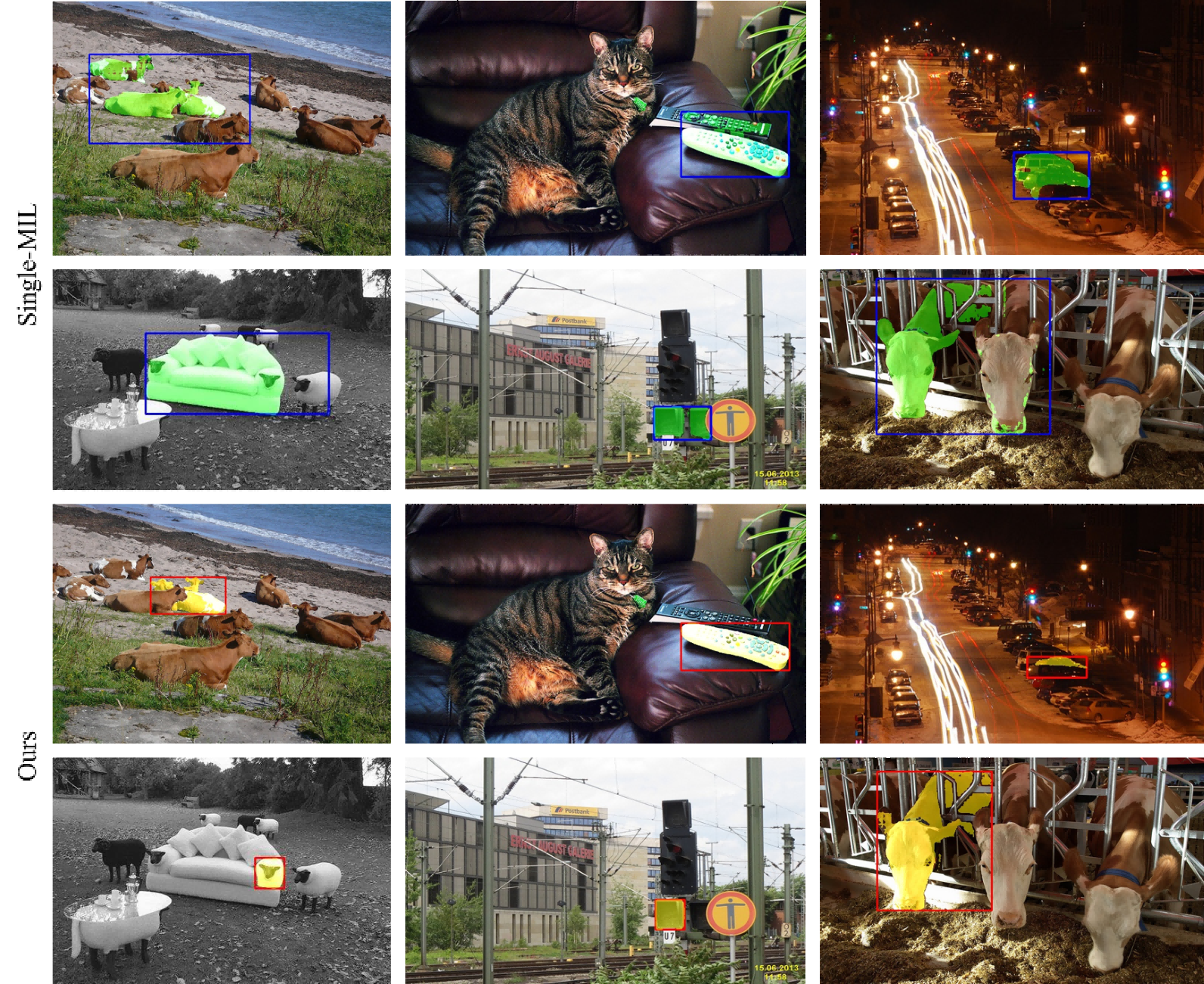}
    \end{tabular}
   \captionsetup{width=0.98\linewidth
   } 
   \caption{Visualization of the segmentation problem for similar neighboring objects, showing single-MIL's outcomes in green masks and our results in yellow masks. The blue and red bounding boxes highlight the respective mask boundaries.
   }
\label{fig:visual2}
\end{center}
\end{figure*}

To illustrate the advantages of SAPNet over the single-MIL approach, Fig.~\ref{fig:visual0} and Fig.~\ref{fig:visual2} contrast the outcomes of the local segmentation problems and the scenario involving proximate objects of the same class, respectively. With the chosen masks encompassing the entirety of the objects, post-selection and refinement via SAPNet markedly alleviate the local segmentation issues. Furthermore, by incorporating point distance guidance, SAPNet achieves commendable segmentation results even with the adjacent objects of the same class, successfully isolating the masks pertinent to each specific object.

\begin{figure*}[ht]
\begin{center}
    \begin{tabular}{ccc}
    \includegraphics[width=0.98\linewidth]{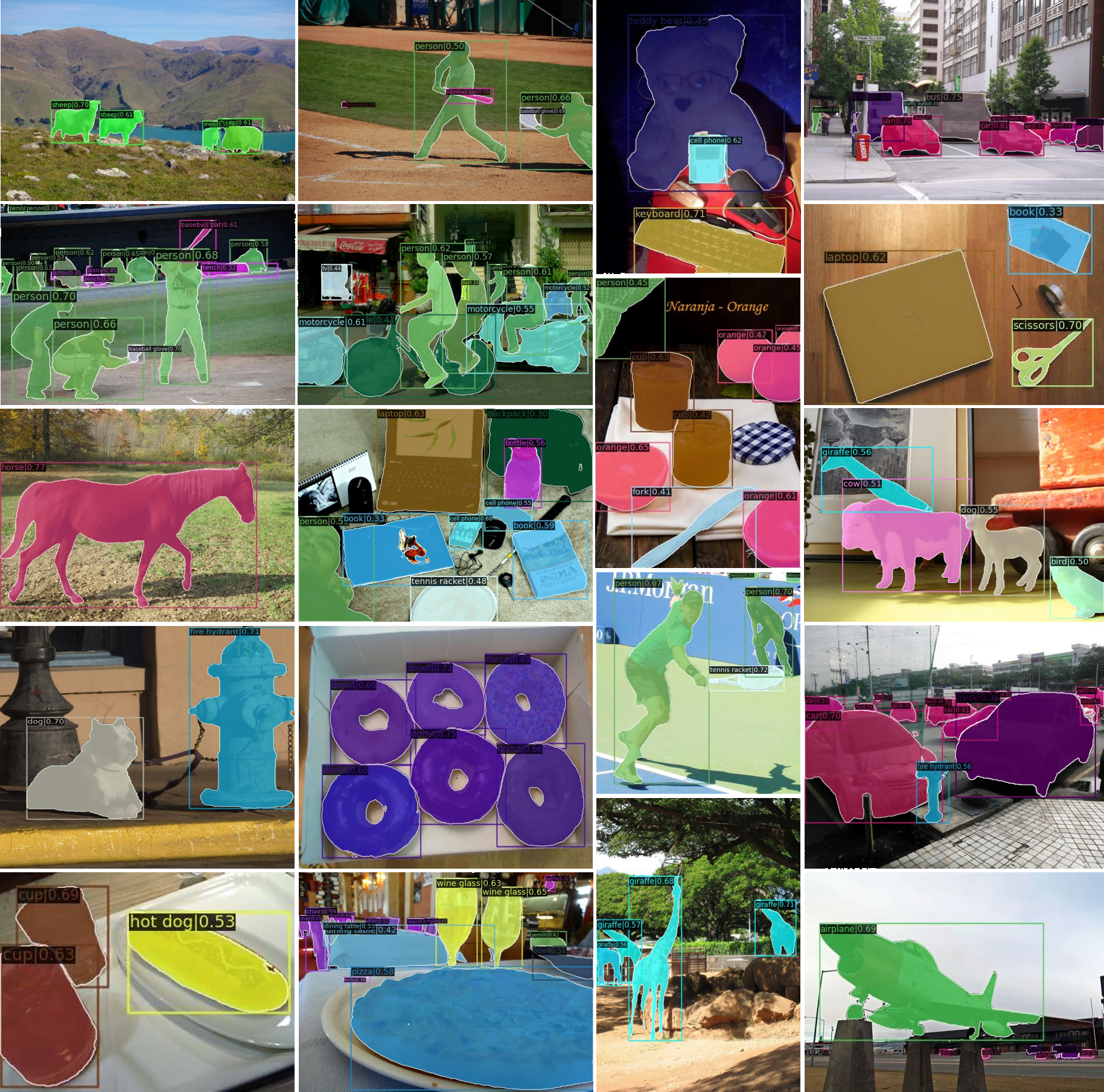}
    \end{tabular}
    \captionsetup{width=0.98\linewidth}
    \caption{Visualization of instance segmentation results utilizing the Resnet-101-FPN backbone. The model is
trained on the COCO train2017 dataset.
   }
\label{fig:visual_our}
\end{center}
\end{figure*}

Fig.~\ref{fig:visual_our} shows additional instance segmentation results of SAPNet on the COCO dataset, demonstrating superior segmentation performance both in the case of individual large objects and in denser scenes. Our method exhibits outstanding capabilities in segmenting singular large targets as well as operating effectively in complex, crowded environments.

\textbf{Ablation studies for negative proposals.}
In Tab.~\ref{tab:NPG threshold}, we evaluate the effect of different threshold settings on SAPNet's final segmentation performance when the negative examples are generated in the NPG on the COCO dataset. We find that the two threshold pairs have less effect on the final segmentation performance with different settings, and the module is robust to hyperparameters.
\begin{table}[!htb]
\begin{center}
    \resizebox{0.48\textwidth}{!}{
\begin{tabular}{c|c|c|c|c|c|c|c}
\hline
$T_{neg1}$ & $T_{neg2}$ & \(AP\) & \(AP_{50}\)& \(AP_{75}\) & \(AP^s\) & \(AP^m\) & \(AP^l\)              \\
\hline
\hline
0.3  &0.3 & 30.9 &   51.3   & 32.0 & 12.2 &   34.7   & 47.4\\

0.3  &0.5 & \textbf{31.2} &   \textbf{51.8}   & \textbf{32.3} & \textbf{12.6} &   \textbf{35.1}   & \textbf{47.8}\\

0.5  &0.3 & 30.8 &   51.1   & 32.0 & 12.1 &   34.7   & 47.3\\

0.5  &0.5 & 31.1 &   51.8   & 32.4 & 12.4 &   35.2   & 47.4\\
\hline
\end{tabular}}
\end{center}
\vspace{-15pt}
\caption{ Constraints in negative proposals generation.}
\vspace{-10pt}
\label{tab:NPG threshold}
\end{table}
\begin{algorithm}[tb!]
{
\small
\caption{Box  Mining Strategy}
\label{Alg: box  mining strategy (bms)}
\textbf{Input:} $T_{min1}$,$T_{min1}$,$box_{select}$from PRM stage,initial positive bag$B^+$,image $I$. \\
\textbf{Output:} Final bounding box for PRM $box_{prm}$.\\
\vspace{-10pt}
\begin{algorithmic}[2]
\FOR{$i \in N$,$N$ is the number of object in image $I$}
\STATE ${G_i} \leftarrow select(B_i,top_k)$ ,$B_i \in B^+$;
\STATE $count = 0$;
\IF {$proposal_j^w \cdot proposal_j^h > box_{select}^h \cdot  box_{select}^w$ \STATE for each $proposal_j \in G_i$\\}
\STATE $iou = IOU(proposal_j, box_{select})$;\\
\IF{ $iou > T_{min1}$ } 
\STATE $box_{prm}$ = \\
\STATE${(proposal_{j} + iou \cdot box_{select})}/{(iou+1)}$;
\STATE $T_{min1} = iou$
\STATE $count+=1$;
\ELSIF {{$count==0$ and \\
\STATE \ \ \ \ \ \ \ \ \ \ $box\_{select} \in proposal_{j}$ 
\STATE and\\
\STATE \ \ \ \ \ \ \ \ \ \ $iou > T_{min2}$\\ }}
\STATE $box_{prm}$ = 
\STATE ${(proposal_{j} \cdot iou + box_{select})}/{(iou+1)}$;
\STATE $T_{min2} = iou$
\ENDIF
\ENDIF
\ENDFOR
\end{algorithmic}
}
\end{algorithm}

\section{Detection and segmentation performance of SAPNet}
\label{sec:det and seg}
\textbf{Detection performance on COCO17 and VOC2012SBD.} As shown in Tab.~\ref{tab:COCO_VOC_DET}, utilizing the COCO and VOC datasets, we conduct an extensive comparison of the proposed methodology against a range of detection methods, encompassing fully, image-level, and point annotation. On the COCO dataset, our method demonstrates a notable improvement over the current SOTA P2BNet \cite{DBLP:cpf}, with a substantial increment of 10.4 $AP$ and culminating in a score of (32.5 $AP$ \emph{vs} 22.1 $AP$). Moreover, under a training schedule extended to 3x, the detection efficacy of SAPNet equates to that of the fully-annotated FPN \cite{DBLP:FPN}. Also, under a 3x training schedule, within the detection of the VOC dataset, our approach exceeds the previous SOTA by 8.0 $AP_{50}$, approximating $91\%$ efficacy of the fully-annotated FPN. We observe that the image-level methods significantly underperform that of the point-annotated methods on the challenging COCO dataset, achieving only $36\%$ of the performance of the fully-annotated approaches. That distinctly accentuates the advantageous that point-prompted methods, optimizing the trade-off between the economy of annotation efforts and the robustness of detection performance.
\begin{table}[H]
    \centering
    \resizebox{1.\linewidth}{!}{
    \begin{tabular}{l|cccc}
    \specialrule{0.13em}{0pt}{0pt}
         Method&\makebox[0.025\textwidth][c]{Sup.}&\makebox[0.02\textwidth][c]{BB.}&\makebox[0.125\textwidth][c]{COCO17($AP$)}&\makebox[0.11\textwidth][c]{VOC12($AP_{50}$)}\\
             \specialrule{0.08em}{0pt}{0pt}
        FPN~\cite{DBLP:FPN} &$\mathcal{B}$&R-50&37.4&75.3\\
        CASD~\cite{DBLP:CASD} &$\mathcal{I}$&VGG-16&12.8&53.6\\
        CASD~\cite{DBLP:CASD} &$\mathcal{I}$&R-50&13.9&56.8\\
        OD-WSCL~\cite{DBLP:OD-WSCL} & $\mathcal{I}$&VGG-16&13.6&56.2\\
        OD-WSCL~\cite{DBLP:OD-WSCL} & $\mathcal{I}$&R-101&14.4&-\\
        UFO$^2$~\cite{DBLP:ufo2}&$\mathcal{P}$&VGG-16&13.0&41.0\\
        UFO$^{2{\ddagger}}$~\cite{DBLP:ufo2}&$\mathcal{P}$&R-50&13.2&38.6 \\
        P2BNet-FR~\cite{DBLP:cpf}&$\mathcal{P}$&R-50&22.1&48.3\\
\hline
\hline
        \rowcolor[rgb]{ 0.902,  0.902,  0.902}\textbf{SAPNet(\emph{ours})}&$\mathcal{P}$&R-50& \textbf{32.5} &\textbf{64.8}\\
        \rowcolor[rgb]{ 0.902,  0.902,  0.902}\textbf{SAPNet(\emph{ours})}$^{*}$&$\mathcal{P}$&R-101& \textbf{37.2} &\textbf{68.5}\\
        \specialrule{0.13em}{0pt}{0pt}
    \end{tabular}}
    \caption{ Object detection performance on COCO 2017 and VOC 2012 $test$ sets. The evaluation metric are $\rm AP$ and $\rm AP_{50}$ respectively.}
    \vspace{-10pt}
    \label{tab:COCO_VOC_DET}
\end{table}
\vspace{-0.8cm}
%